\documentclass[3p,times,twocolumn,numafflabel]{elsarticle}

\usepackage{amsmath,amssymb}
\usepackage{booktabs}
\usepackage{graphicx}
\usepackage{hyperref}
\usepackage{algorithm}
\usepackage{algpseudocode}
\usepackage{xspace}
\usepackage{cuted}
\hypersetup{hypertexnames=false}
\graphicspath{{figures/}{../figures/}}
\makeatletter
\long\def\@makefntext#1{\noindent\parbox[t]{\columnwidth}{\raggedright #1}}
\makeatother
\journal{Infectious Medicine}

\begin{document}

\begin{frontmatter}
  \title{Ensemble Deep Learning Models for Early Detection of Meningitis in ICU: Multi-center Study}
  \author[aff1]{Han Ouyang}
  \ead{ouyang.ha@northeastern.edu}
  \author[aff4]{Ayush Singhal}
  \ead{singha50@purdue.edu}
  \author[aff2]{Jesse Hamilton}
  \ead{hamilton.je@northeastern.edu}
  \author[aff3]{Saeed Amal\corref{cor1}}
  \ead{s.amal@northeastern.edu}
  \cortext[cor1]{*Corresponding author: Saeed Amal}
  \affiliation[aff1]{organization={Khoury College of Computer Science, The Roux Institute, Northeastern University}, city={Portland}, state={ME}, postcode={04101}, country={USA}}
  \affiliation[aff3]{organization={The Roux Institute, Department of Bioengineering, College of Engineering, Northeastern University}, city={Boston}, state={MA}, postcode={02115}, country={USA}}
  \affiliation[aff4]{organization={Computer Science Department, College of Science, Purdue University}, city={West Lafayette}, state={IN}, postcode={47906}, country={USA}}
  \vspace{-0.8em}
  \begin{abstract}
    \textbf{Background}: Meningitis is an inflammatory condition affecting the meninges, which may lead to neurological injury and death. The symptoms are similar to other severe diseases. It can take up to 48 hours for cerebrospinal fluid (CSF) culture to confirm a diagnosis of meningitis. However, advancement of machine learning (ML) with electronic health records (EHRs) has provided researchers and clinicians an innovative approach of detecting meningitis in the settings of intensive care unit (ICU) and emergency room (ER).\\
\textbf{Methods}: 214 meningitis patients (ICD-9: 322.x) and 46,303 non-meningitis patients were identified from the MIMIC-III v1.4 dataset. After temporal filtering, 5,580 samples with 6,970 ICD-9 and demographic variables were extracted. We trained three base classifiers (Random Forest (RF), LightGBM, and a Deep Neural Network (DNN)) with cost-sensitive class weighting, and the outputs were aggregated by a logistic regression meta-learner. The performance of the ensemble classifier was evaluated on three different datasets including a balanced internal test set ($n = 68$), an imbalanced internal test set ($n = 7{,}382$), and an external eICU cohort ($n = 27{,}156$), as well as measured through DeLong tests and SHapley Additive exPlanations (SHAP).\\
\textbf{Results}: In the balanced test set, the ensemble model yielded an area under the receiver operating characteristic curve (AUC) of 0.9424 (95\% CI: 0.8840–0.9870) with a sensitivity of 0.8514. The AUC of the imbalanced test set ($n = 7{,}382$) was 0.9305, with a negative predictive value (NPV) of 0.9993. The stacking ensemble provided statistically significant AUC superiority over both LightGBM ($p = 0.0002$) and DNN ($p = 0.0010$) via DeLong tests. The AUC of the external eICU cohort ($n = 27{,}156$) was 0.6855 with a sensitivity of 0.8455.\\
\textbf{Conclusions}: The stacking ensemble combining RF, LightGBM, and DNN performed well on internal test sets, exhibiting an NPV greater than 99.9\% even with substantial class imbalance. While performance was lower on the external eICU cohort compared to the internal test sets, sensitivity remained robust. Therefore, the stacking ensemble may serve as a rule-out screening option for ERs and ICUs after additional prospective multi-site validation studies for its efficacy in real-world. 
  \end{abstract}
  \vspace{-0.6em}
  \begin{keyword}
    meningitis \sep ensemble learning \sep stacking \sep EHR \sep MIMIC-III \sep imbalanced classification\\
    Graphical abstract: A visual summary is provided in Fig.~\ref{fig:graphical-abstract}.
  \end{keyword}
\end{frontmatter}

\begin{figure*}[t]
\centering
\includegraphics[width=0.95\textwidth]{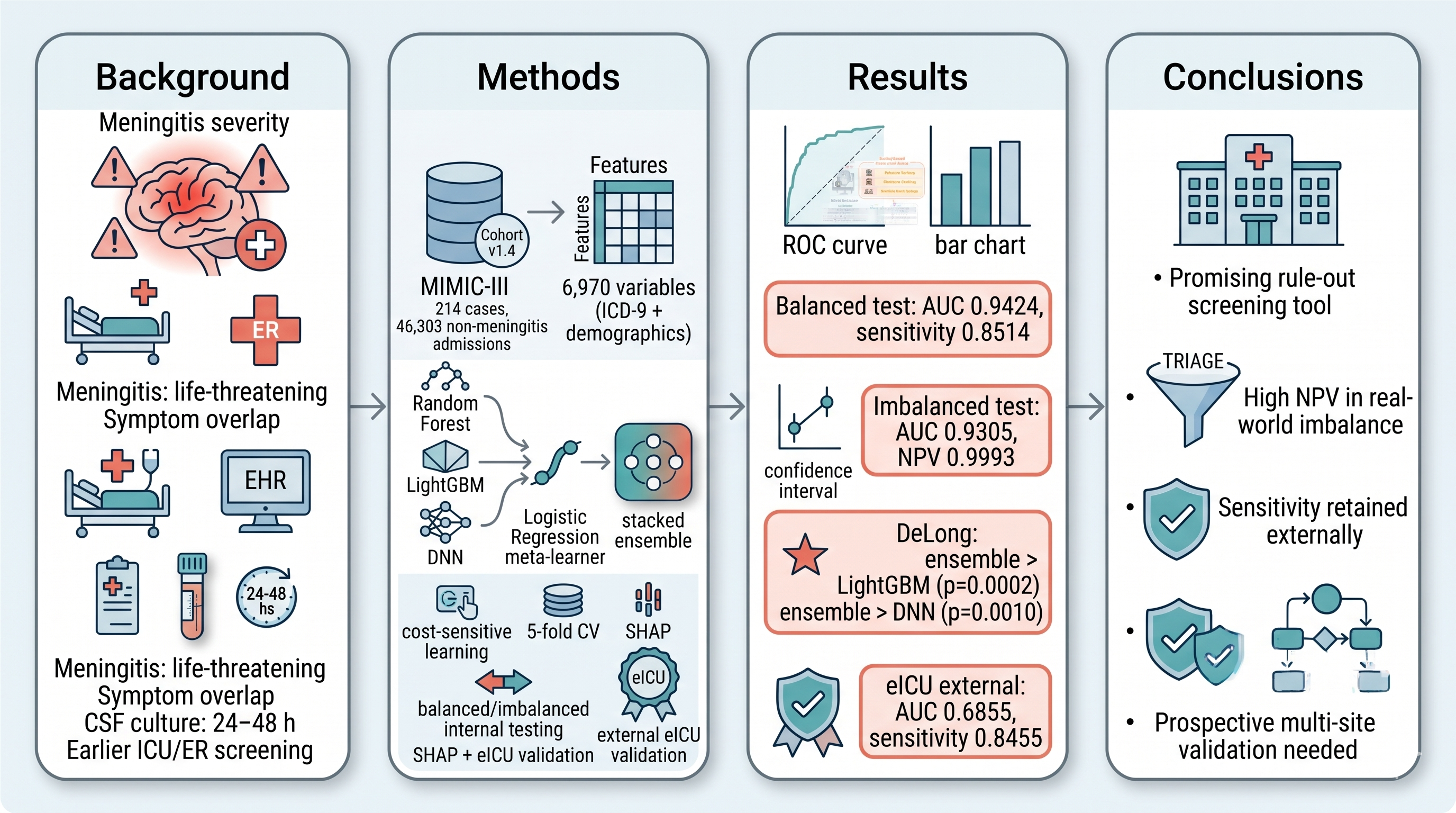}
\caption{Graphical abstract summarizing the multi-center stacking framework for early meningitis detection in ICU cohorts.}
\label{fig:graphical-abstract}
\end{figure*}

\section{Introduction}

Meningitis is a serious inflammatory condition affecting the meninges around the brain and spinal cord \cite{ref1,ref2}. Early diagnosis is important because delays can lead to neurological injury or death \cite{ref3,ref4,ref5}. In ER and ICU settings, however, diagnosis can be challenging because the symptoms are often nonspecific or resemble those of other conditions \cite{ref6,ref7}. Bacterial meningitis is diagnosed using cerebrospinal fluid culture (the "gold standard"), which may take up to 48 hours to determine the causative organism, and during this time, health care professionals must make empiric treatment decisions without knowing the pathogen. Due to the delayed diagnosis, it is worth exploring the potential of EHR systems to develop machine learning models which are able to identify severe cases.

EHRs provide structured patient data at scale, making them well suited for predictive modeling in healthcare \cite{ref8,ref39} and for detecting conditions that are frequently missed or misdiagnosed \cite{ref9,ref10,ref11}. Ensemble and deep learning methods have demonstrated strong performance across various medical domains, including detection, screening, classification, identification, diagnosis, and explainability for clinical trust \cite{ref40,ref41,ref50,ref51,ref52,ref53,ref54,ref55,ref56,ref57,ref58,ref59}. 

Meningitis was rare in the MIMIC-III cohort, accounting for only 0.46\% of ICU admissions (214 of 46{,}517). This substantial class imbalance made conventional modeling strategies less suitable and motivated the use of imbalance-aware methods \cite{ref12,ref13,ref14}. To address this issue, we utilized cost-sensitive learning with undersampling during model training \cite{ref15,ref16,ref17}. Three base models (RF, LightGBM, and DNN) were trained using 5-fold cross-validation. Their predicted probabilities were then aggregated using a logistic regression meta-learner to build a stacking ensemble \cite{ref18,ref19,ref20}.

\begin{figure*}[t]
  \centering
  \includegraphics[width=0.92\textwidth]{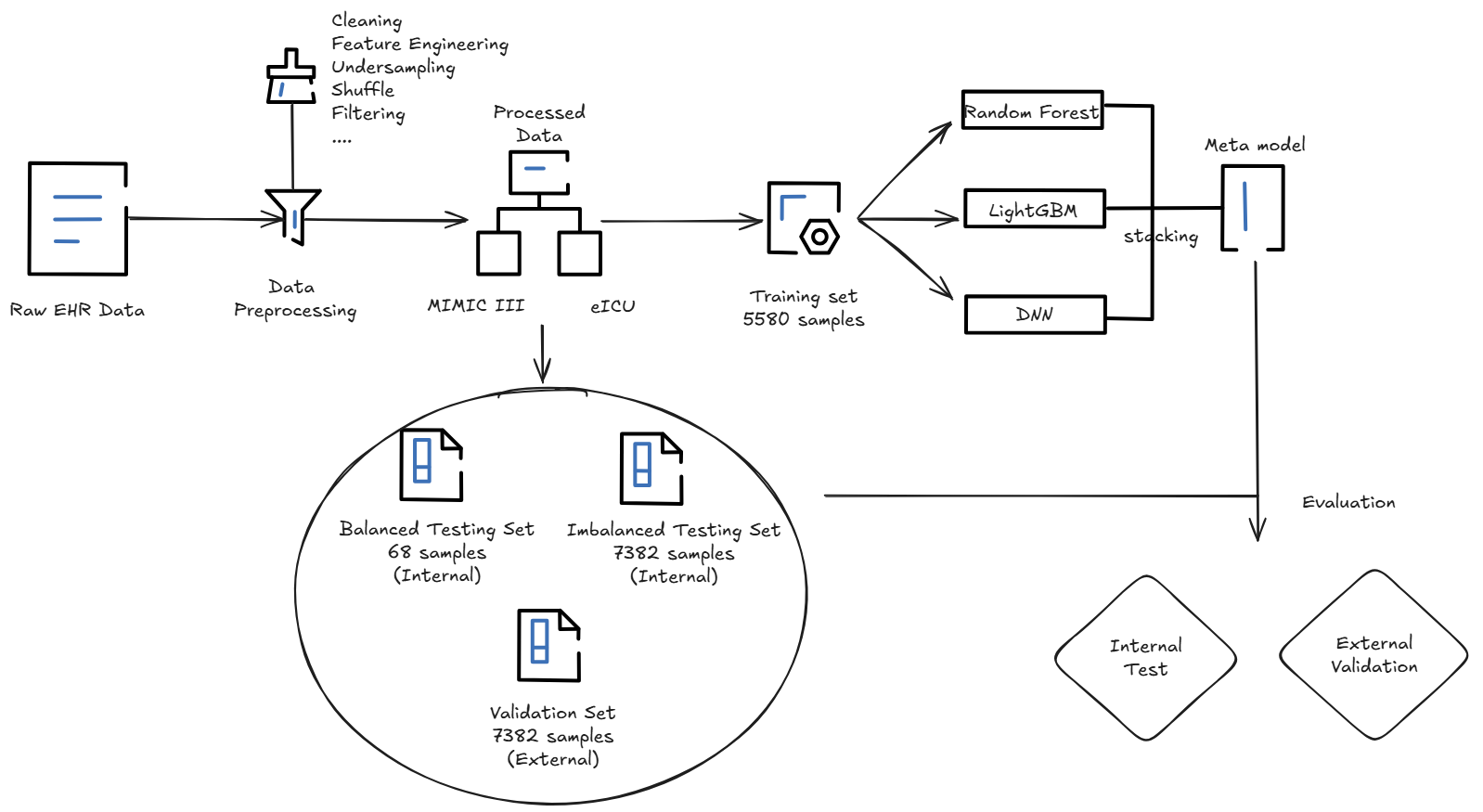}
  \caption{Overview of the proposed meningitis prediction workflow. Raw EHR data undergo cleaning, feature engineering, undersampling, shuffling, and filtering before cohort construction from MIMIC-III and eICU. A stacking ensemble combining Random Forest, LightGBM, and DNN is trained on the MIMIC-III training set and evaluated on balanced and imbalanced internal test sets, followed by external validation on eICU.}
  \label{fig:intro_workflow}
\end{figure*}

Model performance was evaluated in three settings: a balanced test set ($n = 68$; 50\% positives), an imbalanced test set with more realistic ICU prevalence ($n = 7{,}382$; 0.46\% positives), and an external eICU cohort ($n = 27{,}156$; 0.46\% positives). We assessed performance using AUC, sensitivity, specificity, PPV, NPV, F1-score, MCC, and PR-AUC, each reported with 95\% bootstrap confidence intervals. DeLong tests were used to compare the AUC of the ensemble with that of each base model, and model-specific decision thresholds were determined using Youden's J statistic. An overview of the study workflow is shown in Fig.~\ref{fig:intro_workflow}.

Prior ML studies of meningitis detection have reported AUC of 0.83 \cite{ref21}, accuracy of 96.69\% \cite{ref22}, AUC of 0.930 \cite{ref24}, and sensitivity of 83.3\% \cite{ref25}; one study used regression without modern ML algorithms \cite{ref23}. Our work extends these by using a larger realistic dataset, a three-model stacking ensemble with SHAP-based interpretability \cite{ref34,ref36}, and evaluation under both balanced and clinically imbalanced conditions with external validation.

This study has three main contributions. First, we show that a stacking ensemble can improve rare-disease prediction under severe class imbalance. Second, we examine whether ICD-9 codes can serve as informative structured features and use RF importance and SHAP to identify the factors driving predictions. Third, we evaluate the model on both balanced and imbalanced internal test sets, along with an external eICU cohort, to assess the performance and the capability of the models in terms of generalization.

\section{Materials and Methods}

\subsection{Dataset Overview}

This study utilized data collected from the MIMIC-III \cite{ref43}, a publicly available clinical data warehouse maintained by Massachusetts Institute of Technology (MIT) and Harvard University.  Cases of meningitis were annotated via ICD 9 diagnosis codes beginning with 322 (ICD-9 code: 322.x). There were 214 meningitis cases and 46,303 other ICU admissions, resulting in a total patient cohort of 46,517 patients with an approximate meningitis prevalence of 0.46\%. The demographic features that were included as input variables for the model included gender and ICD-9 codes. To ensure temporal purity, only clinical features documented before the timestamp of the first ICD-9 322.x diagnosis were retained for each meningitis case in order to prevent post-diagnosis information from leakage.

\subsection{Data Preprocessing}

Before the model training, data preparation happened in a preprocessing pipeline that used input files containing pre-processed feature matrices. The preprocessing pipeline contained a series of steps that included reviewing records to find and remove duplicate and inconsistent records; removing irrelevant columns from each record; filtering the ICD-9 codes in each patient's post-diagnosis records (for all records after each patient's first record with an ICD-9 code for 322.x); merging records for each patient based on their unique patient IDs; and finally, zero-imputing the null values in the binary feature matrix for ICD-9 codes, which indicated that there was no diagnostic code given to the patient.

\subsection{Feature Engineering}

To reduce overfitting when creating models based on the meningitis data, we only utilized features that were present prior to the time of diagnosis, including gender and diagnostic codes for previously diagnosed or concurrently diagnosed comorbidities (excluding diagnostic codes assigned after diagnosis). The ICD-9 diagnostic codes were one-hot encoded \cite{ref30,ref31}, producing a sparse binary feature matrix with a high degree of dimensionality. Feature importance, as determined by random forests, was used to display clinical plausibility and statistical significance while mitigating overfitting and leakage.

Gender and ICD-9 diagnostic codes were retained as the model input features. Although additional demographic variables including patient ethnicity, admission type, and admission time were evaluated as candidate features, their inclusion did not improve predictive performance and introduced additional noise into the model according to our multiple experiments; consequently, these variables were excluded. This is consistent with the principle that parsimonious feature representations can outperform more complex ones in clinical prediction tasks with sparse, high-dimensional data. Gender was identified as among the strongest individual features according to RF feature importance (Table~\ref{tab:2}). The datasets were one-hot encoded and combined to yield a full training feature matrix of 6,970 variables.

The top 100 features by cumulative Random Forest importance capture 96\% of total cumulative RF importance (Figure~\ref{fig:2}), providing a principled basis for this selection. These top 100 features are used for the feature importance display (Table~\ref{tab:2}) and SHAP-based interpretability analysis; all base models were trained on the full 6,970-feature matrix.

\subsection{Handling Class Imbalance}

The data exhibited an extreme degree of class imbalance relative to other datasets, with only 0.46\% of subjects in this study diagnosed with meningitis. If the models had been trained on the full dataset, they would likely have produced biased results favoring the majority class \cite{ref12,ref14}. To mitigate this severe class imbalance, we constructed the training set by undersampling the majority class \cite{ref42}. This process resulted in a training set containing 5,580 samples, including 180 meningitis cases and 5,400 non-meningitis controls, corresponding to a ratio of 1:30. All meningitis training examples were retained, while the non-meningitis controls were undersampled. This undersampling strategy helped ensure that both the base models and the ensemble model did not merely reflect the majority class, thereby supporting the maintenance of model sensitivity and specificity \cite{ref32,ref33}.

Generalization is facilitated in two ways. The first method is to utilize the out-of-fold predictions as both the meta-feature matrix and base feature matrix to prevent information leakage so that the meta-model is only trained on unseen data. The second method is to test the model against a large imbalanced test dataset that reflects the real-world prevalence of ICU patients, thereby providing the model with a real-world scenario in which to assess the robustness of the model.

As a complementary mechanism to undersampling, cost-sensitive learning was applied to each base model to further address class imbalance. The Random Forest classifier required the use of a custom class weight dictionary specifying as $\{0: 1.0,\; 1: (n_{\text{neg}}/n_{\text{pos}}) \times 0.8\}$, where $n_{\text{neg}}$ and $n_{\text{pos}}$ denote the number of negative and positive training samples, respectively. For LightGBM, the \texttt{scale\_pos\_weight} hyperparameter was set to $(n_{\text{neg}}/n_{\text{pos}}) \times 0.75$, which scales the gradient contribution of positive-class samples during boosting. For the Deep Neural Network (DNN), a WeightedFocalLoss function was employed with class weights $[1.0,\; n_{\text{neg}}/n_{\text{pos}}]$ and a focusing parameter $\gamma = 2.0$, which down-weights easy negative examples and concentrates learning on the minority meningitis class.Together, undersampling and cost-sensitive weighting ensure sufficient representation of minority-class patterns in the training dataset while not losing the value of the information contained within majority-class instances.

\subsection{Experimental Design}

Three classifiers were built and assessed: (1) Random Forest (RF), a nonparametric ensemble approach that can efficiently classify high-dimensional categorical variables; (2) LightGBM, an optimized gradient-boosting technique that performs well with sparse datasets; and (3) Deep Neural Network (DNN), a multi-layer perceptron-based architecture capable of learning complex nonlinear interactions. All models were implemented using the Python programming language, utilizing the scikit-learn, PyTorch, and LightGBM libraries, and trained with stratified 5-fold cross-validation. Model performance was evaluated using the area under the receiver operating characteristic curve (AUC), sensitivity, specificity, positive predictive value (PPV), negative predictive value (NPV), F1 score, Matthews correlation coefficient (MCC), precision-recall area under the curve (PR-AUC), and 95\% bootstrap confidence intervals (CIs).

During the stacking ensemble phase, the out-of-fold prediction probabilities from all three classifiers were stacked to generate a three-column meta-feature matrix. Each column corresponds to one classifier's predictions \cite{ref35}. A logistic regression model was employed as the meta-learner to learn a linear combination of the predictions provided by the three individual classifiers, based upon the meta-feature matrix \cite{ref35}. The meta-learner was trained solely using the predictions for the corresponding fold that the individual classifier did not predict, thus maintaining the integrity of the cross-validation process and preventing the potential leakage of information \cite{ref37}.

The evaluation of the model utilized two sets of internal testing. The balanced testing set ($n = 68$; 34 cases of meningitis and 34 without) served as the primary performance evaluation under balanced conditions. The imbalanced test set (n = 7,382; 34 meningitis cases and 7,348 non-meningitis) reflects the real-life incidence of meningitis in an ICU and allows for a clinical-like assessment of the model's behaviour under extreme class imbalance. An external evaluation of the model was completed using the eICU Collaborative Research Database (eICU-CRD) \cite{ref44}. This external dataset included 27,156 patients (124 meningitis cases and 27,032 non-meningitis), with cases identified using the same ICD-9 322.x criterion. Algorithm 1 \ref{alg:1} a summary of the entire pipeline used for training and inference of the model. Upon publication, the full implementation code will be available.

\begin{algorithm}[htbp]
\caption{Stacking Ensemble Training and Inference}
\label{alg:1}
\begin{algorithmic}[1]
\Require Training set $D_{\text{train}}$ (180 meningitis, 5,400 non-meningitis); test sets $D_{\text{bal}}$, $D_{\text{imbal}}$, $D_{\text{eICU}}$
\Ensure Meningitis probability $\hat{p}$ and binary prediction $\hat{y}$ for each patient
\State \textbf{Train base models with 5-fold cross-validation}
\For{$k = 1$ \textbf{to} $5$}
  \State Split $D_{\text{train}}$ into fold $D_k^{\text{trn}}$ and $D_k^{\text{val}}$
  \State Train RF, LightGBM, DNN on $D_k^{\text{trn}}$ with cost-sensitive class weights
  \State Generate out-of-fold predictions $p_{\text{RF}}^{(k)}$, $p_{\text{LGB}}^{(k)}$, $p_{\text{DNN}}^{(k)}$ on $D_k^{\text{val}}$
\EndFor
\State Stack predictions: $Z_{\text{train}} \leftarrow [p_{\text{RF}},\; p_{\text{LGB}},\; p_{\text{DNN}}]$ --- an $N_{\text{train}} \times 3$ meta-feature matrix
\State \textbf{Train meta-learner}
\State Fit logistic regression (L2) on $Z_{\text{train}}$ with meningitis labels $y$
\State Learn coefficients $\beta_{\text{RF}},\; \beta_{\text{LGB}},\; \beta_{\text{DNN}}$
\State \textbf{Retrain base models on full $D_{\text{train}}$}
\For{each test set $D_{\text{test}} \in \{D_{\text{bal}},\; D_{\text{imbal}},\; D_{\text{eICU}}\}$}
  \State Generate base predictions $p_{\text{RF}},\; p_{\text{LGB}},\; p_{\text{DNN}}$ on $D_{\text{test}}$
  \State Stack: $Z_{\text{test}} \leftarrow [p_{\text{RF}},\; p_{\text{LGB}},\; p_{\text{DNN}}]$
  \State Compute ensemble probability: $\hat{p} = \sigma(\beta_0 + \beta_{\text{RF}} p_{\text{RF}} + \beta_{\text{LGB}} p_{\text{LGB}} + \beta_{\text{DNN}} p_{\text{DNN}})$
  \State Apply Youden's J threshold: $\hat{y} = \mathbf{1}[\hat{p} \geq t^*]$
\EndFor
\State \textbf{Evaluate} using AUC, sensitivity, specificity, PPV, NPV, F1, MCC, PR-AUC with 95\% bootstrap CIs; apply DeLong test against each base model
\end{algorithmic}
\end{algorithm}

\subsubsection{Performance Metrics}

In the formulas below, TP denotes true positives; TN denotes true negatives; FP denotes false positives ; FN denotes false negatives. Sensitivity (recall) represents the proportion of meningitis cases correctly detected among real positive cases:

\begin{equation}
\text{Sensitivity} = \frac{TP}{TP + FN}
\end{equation}

Specificity measures the proportion of non-meningitis cases correctly excluded:

\begin{equation}
\text{Specificity} = \frac{TN}{TN + FP}
\end{equation}

Positive predictive value (PPV) is the probability that a positive prediction corresponds to a true meningitis case:

\begin{equation}
\text{PPV} = \frac{TP}{TP + FP}
\end{equation}

Negative predictive value (NPV) is the probability that a negative prediction corresponds to a true non-meningitis case, which is particularly important in high-throughput screening contexts:

\begin{equation}
\text{NPV} = \frac{TN}{TN + FN}
\end{equation}

The F1-score is the harmonic mean of PPV and sensitivity, balancing precision and recall under class imbalance:

\begin{equation}
\text{F1} = \frac{2 \cdot TP}{2 \cdot TP + FP + FN}
\end{equation}

In this severely imbalanced cohort, F1 penalizes both missed meningitis cases (FN) and false alarms (FP) simultaneously.

The Matthews Correlation Coefficient (MCC) provides a single balanced measure across all four confusion matrix cells and is robust to class imbalance:

\begin{equation}
\text{MCC} = \frac{TP \cdot TN - FP \cdot FN}{\sqrt{(TP+FP)(TP+FN)(TN+FP)(TN+FN)}}
\end{equation}

MCC is preferred here because accuracy is a misleading metric at 0.46\% prevalence, where a trivial classifier predicting all non-meningitis patients would yield an accuracy of over 99\% while missing all actual cases of meningitis. The AUC value provides a summary of how well a given model can differentiate between two classes across all possible decision thresholds and is estimated using the nonparametric Wilcoxon–Mann–Whitney statistic. Equivalently, AUC provides a way to quantify the probability that a model will assign a higher risk score for meningitis to a true meningitis patient than to a randomly chosen non-meningitis patient. PR-AUC measures performance given an imbalance of classes and combines positive predictive value and sensitivity to summarize how models perform across all thresholds with respect to predictive power for the rare positive class (meningitis)

\subsubsection{Statistical Analysis and Interpretability}

\textbf{Optimal threshold selection.} Per-model decision thresholds were selected by maximizing Youden's J statistic over the receiver operating characteristic curve:

\begin{equation}
t^* = \underset{t}{\arg\max}\;\left[\text{Sensitivity}(t) + \text{Specificity}(t) - 1\right]
\end{equation}

where $t$ is the predicted meningitis probability threshold. This criterion identifies the operating point that simultaneously maximizes sensitivity and specificity for each model.

\textbf{Bootstrap confidence intervals.} All performance metrics were reported with 95\% bootstrap confidence intervals. For a metric $\hat{\theta}$ computed on a test set of size $N$, $B = 1{,}000$ bootstrap resamples (sampling with replacement) were drawn and $\hat{\theta}^*_b$ was computed for each resample $b$. The 95\% CI was taken as the 2.5th and 97.5th empirical percentiles of $\{\hat{\theta}^*_1, \ldots, \hat{\theta}^*_B\}$. Bootstrap CIs are used because the small number of meningitis test cases ($n = 34$ per internal test set) makes asymptotic normality unreliable.

\textbf{DeLong test.} To evaluate whether ensemble AUC was significantly better than each base model’s AUC on the imbalanced testing set, we used the DeLong test for correlated receiver operating characteristic curves \cite{ref45}. The test statistic is:

\begin{equation}
\resizebox{0.94\columnwidth}{!}{$\displaystyle z = \frac{\widehat{AUC}_{\text{ensemble}} - \widehat{AUC}_{\text{base}}}{\sqrt{\widehat{\mathrm{Var}}(\widehat{AUC}_{\text{ensemble}}) + \widehat{\mathrm{Var}}(\widehat{AUC}_{\text{base}}) - 2\,\widehat{\mathrm{Cov}}(\widehat{AUC}_{\text{ensemble}},\,\widehat{AUC}_{\text{base}})}}$}
\end{equation}

where the variance and covariance are estimated using the DeLong structural component method, with correlation arising due to shared test samples. Statistically significant results indicate that the ensemble has greater discriminative performance than the base model on the imbalanced testing set.

\textbf{Brier score.} We measured calibration according to the Brier score, the mean squared error between predicted meningitis probabilities and observed binary outcomes:

\begin{equation}
\text{BS} = \frac{1}{N} \sum_{i=1}^{N} \left(\hat{p}_i - y_i\right)^2
\end{equation}

where $\hat{p}_i$ is the ensemble's predicted probability of meningitis for patient $i$ and $y_i \in \{0, 1\}$ is the true label. A lower Brier score indicates better calibration; a perfectly calibrated model achieves BS = 0.

\textbf{SHAP interpretability.} SHapley Additive exPlanations (SHAP) were applied to the full stacking ensemble using KernelExplainer. The Shapley value for feature $j$ with respect to input $\mathbf{x}$ is:

\begin{equation}
\phi_j(\mathbf{x}) = \sum_{S \subseteq F \setminus \{j\}} \frac{|S|!\,(|F|-|S|-1)!}{|F|!}\left[f\!\left(\mathbf{x}_{S \cup \{j\}}\right) - f\!\left(\mathbf{x}_S\right)\right]
\end{equation}

Here, $F$ refers to the full set of 6,970 input features, $S$ denotes a feature subset, $F$ represents the ensemble’s \texttt{predict\_proba} function $((\mathrm{RF} + \mathrm{LightGBM} + \mathrm{DNN}) \rightarrow \text{Logistic Regression meta-learner})$, and $\mathbf{x}_S$ denotes the input in which features outside $S$ are replaced with background reference values. A background reference set of 50 samples and an explanation set of 300 samples were used. The mean absolute Shapley value, $\bar{\phi}_j = \mathbb{E}[|\phi_j(\mathbf{x})|]$, was used to rank features according to their overall contribution to ensemble predictions of meningitis. A positive $\phi_j$ value for a given feature indicates that it increased the model’s predicted probability of meningitis for that patient, whereas a negative value indicates that it decreased the prediction.

\subsection{Model Implementations}

\subsubsection{Random Forest}

A RF classifier \cite{ref26,ref27} was implemented using scikit-learn's \texttt{RandomForestClassifier} with \texttt{n\_estimators = 500}, \texttt{n\_jobs = -1} (parallelized CPU computation), as well as Stratified 5-fold cross-validation was applied:

\begin{equation}
\hat{P}(y=1 \mid x) = \frac{1}{T} \sum_{t=1}^{T} h_t(x)
\end{equation}

where $h_t(x) \in [0,1]$ is the class-probability estimate from tree $t$, $T = 500$, and $\hat{P}(y=1 \mid x)$ is the ensemble's estimated probability that patient $x$ has meningitis. Binary predictions are produced by thresholding at the per-model Youden's J optimal threshold.

\subsubsection{LightGBM}

A LightGBM classifier \cite{ref28} was trained using the binary cross-entropy objective:

\begin{equation}
\mathcal{L} = -\frac{1}{N} \sum_{i=1}^{N} \left[ y_i \log(p_i) + (1 - y_i) \log(1 - p_i) \right]
\end{equation}

Here $y_i \in \{0,1\}$ is the meningitis label and $p_i$ is the model's predicted probability of meningitis for patient $i$. The model iteratively adds trees to minimize $\mathcal{L}$, with the final prediction for sample $X_i$ given by $\hat{y}_i = \sum_{t=1}^{T} f_t(X_i)$ and probability output $p_i = \sigma(\hat{y}_i)$. Histogram-based learning and leaf-wise growth reduce memory usage and training time. Stratified 5-fold cross-validation was applied.

\subsubsection{Deep Neural Network}

The DNN \cite{ref29} was implemented in PyTorch with the following architecture: Linear(input $\rightarrow$ 256) + BatchNorm + ReLU + Dropout(0.4), Linear(256 $\rightarrow$ 128) + BatchNorm + ReLU + Dropout(0.3), Linear(128 $\rightarrow$ 64) + ReLU, Linear(64 $\rightarrow$ 2). The model was trained on a graphics processing unit (GPU). A WeightedFocalLoss function was used as the loss criterion, defined as:

\begin{equation}
\text{FL}(p_t) = -\alpha_t\,(1 - p_t)^{\gamma}\,\log(p_t)
\end{equation}

where $p_t$ is the model's predicted probability for the true class, $\alpha_t$ is the per-class weight ($[1.0,\; n_{\text{neg}}/n_{\text{pos}}]$ for non-meningitis and meningitis, respectively), and $\gamma = 2.0$ is the focusing parameter that down-weights easy negative examples and concentrates learning on the minority meningitis class. The Adam optimizer was applied with standard momentum parameters. Stratified 5-fold cross-validation was applied.

\subsubsection{Logistic Regression Meta-Learner}

The meta-learner was built on top of logistic regression, which was trained on stacked out-of-fold probability estimates from the three base models:

\begin{equation}
P(y=1 \mid \mathbf{z}) = \sigma\!\left(\beta_0 + \beta_{\text{RF}}\, p_{\text{RF}} + \beta_{\text{LGB}}\, p_{\text{LGB}} + \beta_{\text{DNN}}\, p_{\text{DNN}}\right)
\end{equation}

The learned coefficients $\beta_{\text{RF}}$, $\beta_{\text{LGB}}$, and $\beta_{\text{DNN}}$ represent the relative contribution of each base model to the ensemble prediction. The coefficients of base models serve as weighted contributors to the final ensemble. At the inference stage, the calibrated probability score was outputted by the meta-learner with a threshold based on the optimal value of Youden's J.

\textbf{Training times.} We present the training times from the main experiment (i.e., Train = 5,580 samples) as shown in Table~\ref{tab:1}.

\begin{table}[htbp]
\caption{Model training times (Train = 5,580 samples).}
\label{tab:1}
\centering
\scriptsize
\setlength{\tabcolsep}{3pt}
\resizebox{\columnwidth}{!}{%
\begin{tabular}{@{}lcc@{}}
\toprule
\textbf{Model} & \textbf{Training Samples} & \textbf{Training Time (s)} \\
\midrule
RF & 5,580 & 22.6 \\
LightGBM & 5,580 & 6.9 \\
DNN & 5,580 & 115.3 \\
Logistic Regression (LR, meta) & --- & Negligible \\
\bottomrule
\end{tabular}%
}
\end{table}

\section{Results}

\subsection{Risk Feature Analysis}
\begin{figure*}[t]
\centering
\includegraphics[width=\textwidth]{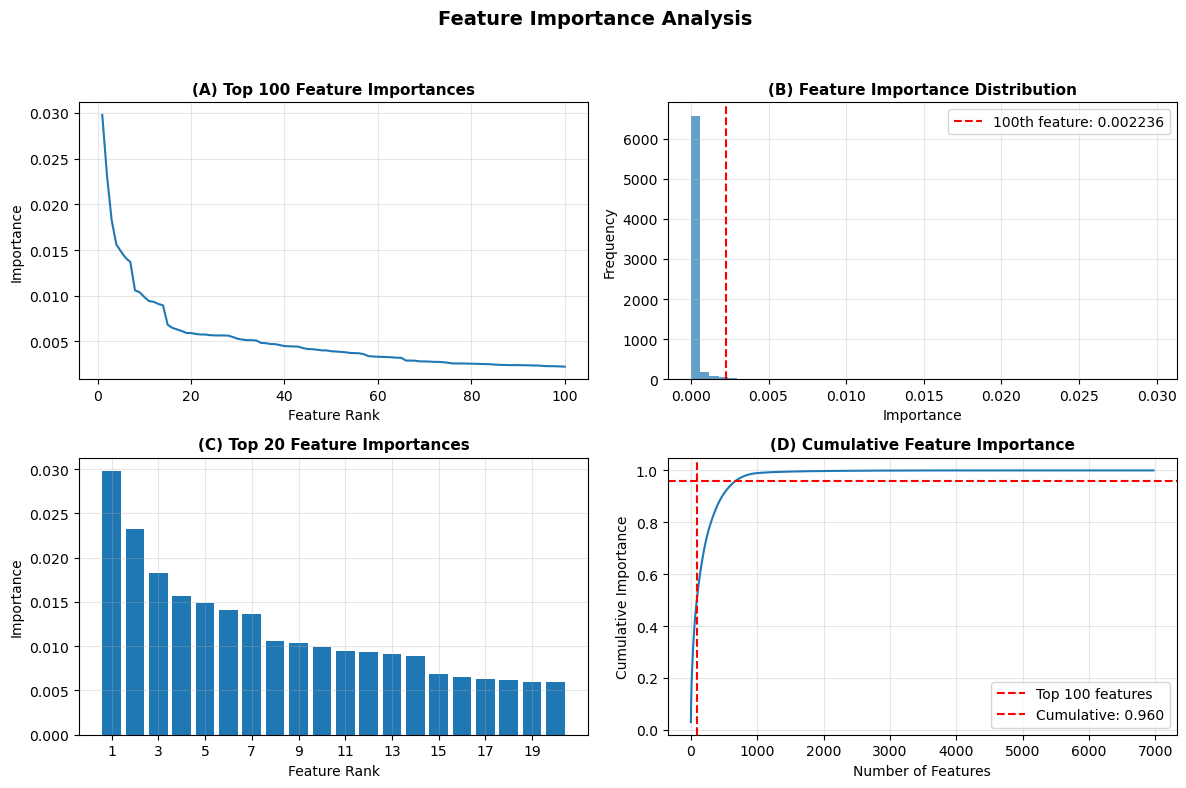}
\caption{Random Forest Cumulative Feature Importance Curve. Cumulative normalized importance plotted against feature rank for all 6,970 variables. Red dashed lines indicate the top-100 feature cutoff and corresponding cumulative importance value. The top 100 features account for 96\% of total RF importance.}
\label{fig:2}
\end{figure*}

Feature importance was processed to examine 6,970 ICD-9 and demographic variables using a RF classifier trained with the whole samples. The top 20 most important features are included in Table~\ref{tab:2}. Figure~\ref{fig:2} shows the cumulative RF importance curves for the top 100 features.

\begin{table}[htbp]
\caption{Top 20 Risk Features Identified by Random Forest Importance. ICD-9 clinical descriptions are provided for each code; GENDER is the only demographic variable included.}
\label{tab:2}
\centering
\scriptsize
\setlength{\tabcolsep}{3pt}
\resizebox{\columnwidth}{!}{%
\begin{tabular}{@{}clp{0.48\columnwidth}c@{}}
\toprule
\textbf{Rank} & \textbf{ICD-9 Code} & \textbf{Clinical Description} & \textbf{RF Importance} \\
\midrule
1  & 41401  & Coronary atherosclerosis of native coronary artery     & 0.029788 \\
2  & 3314   & Obstructive hydrocephalus                              & 0.023201 \\
3  & 430    & Subarachnoid hemorrhage                                & 0.018300 \\
4  & 99663  & Infection/inflammatory reaction due to vascular device/implant & 0.015618 \\
5  & 78039  & Other alteration of consciousness                      & 0.014837 \\
6  & 431    & Intracerebral hemorrhage                               & 0.014136 \\
7  & V290   & Observation for suspected infectious condition         & 0.013691 \\
8  & GENDER & Sex (demographic variable)                             & 0.010588 \\
9  & V053   & Need for prophylactic vaccination, viral hepatitis     & 0.010376 \\
10 & 42731  & Atrial fibrillation                                    & 0.009854 \\
11 & 7907   & Bacteremia                                             & 0.009410 \\
12 & 51881  & Acute respiratory failure                              & 0.009336 \\
13 & 4019   & Unspecified essential hypertension                     & 0.009102 \\
14 & 2761   & Hyposmolality and/or hyponatremia                      & 0.008943 \\
15 & 4280   & Congestive heart failure, unspecified                  & 0.006816 \\
16 & 77181  & Septicemia of newborn                                  & 0.006478 \\
17 & 2724   & Other and unspecified hyperlipidemia                   & 0.006309 \\
18 & 5849   & Acute kidney failure, unspecified                      & 0.006135 \\
19 & 769    & Respiratory distress syndrome                          & 0.005922 \\
20 & 3453   & Grand mal status epilepticus                           & 0.005913 \\
\bottomrule
\end{tabular}%
}
\end{table}

\subsection{Base Model Performance}
Table~\ref{tab:3} provides 5-fold stratified cross-validation outputs from base models with the training dataset ($n = 5{,}580$; 180 meningitis : 5,400 non-meningitis; 1:30 class ratio). To handle class imbalance, all three models were trained with cost-sensitive class weighting. Random Forest obtained the highest AUC value of 0.8784 (95\% CI 0.8495 to 0.9057), sensitivity of 0.7778 (95\% CI 0.7222 to 0.8583), and specificity of 0.8876 (95\% CI 0.8592 to 0.9221). LightGBM achieved an AUC of 0.8228 (95\% CI 0.7687 to 0.8617), sensitivity of 0.7333, and lower specificity of 0.7574 than that of Random Forest. DNN reached an AUC of 0.8242 (95\% CI 0.7644 to 0.8782) with sensitivity of 0.7333 and specificity of 0.8546. NPV was greater than 0.98 on all three base models, suggesting that the models were all able to predict negative outcomes accurately even at a heavily imbalanced condition (1:30). PR-AUC and MCC were higher for RF (0.3714 and 0.3494) than for LightGBM (0.2579 and 0.2029) and DNN (0.2863 and 0.2806). Cross-validation ROC curves for the RF, LightGBM, and DNN models can be found in Supplementary Figure~\ref{fig:s1}.

\begin{table}[htbp]
\caption{Base Model 5-Fold Cross-Validation Performance. Metrics are reported as bootstrap mean (95\% CI, $B = 1{,}000$) on the training set ($n = 5{,}580$; 180 meningitis : 5,400 non-meningitis). Thresholds selected by Youden's J statistic per fold.}
\label{tab:3}
\centering
\scriptsize
\setlength{\tabcolsep}{2pt}
\resizebox{\columnwidth}{!}{%
\begin{tabular}{@{}lccc@{}}
\toprule
\textbf{Metric} & \textbf{RF (95\% CI)} & \textbf{LightGBM (95\% CI)} & \textbf{DNN (95\% CI)} \\
\midrule
AUC        & 0.8784 (0.8495--0.9057) & 0.8228 (0.7687--0.8617) & 0.8242 (0.7644--0.8782) \\
Sensitivity & 0.7778 (0.7222--0.8583) & 0.7333 (0.7222--0.7500) & 0.7333 (0.6694--0.8028) \\
Specificity & 0.8876 (0.8592--0.9221) & 0.7574 (0.6657--0.8313) & 0.8546 (0.8466--0.8743) \\
PPV        & 0.1930 (0.1590--0.2392) & 0.0960 (0.0676--0.1260) & 0.1443 (0.1294--0.1573) \\
NPV        & 0.9917 (0.9897--0.9946) & 0.9883 (0.9863--0.9894) & 0.9897 (0.9872--0.9923) \\
F1-score   & 0.3074 (0.2626--0.3605) & 0.1690 (0.1235--0.2146) & 0.2410 (0.2170--0.2571) \\
MCC        & 0.3494 (0.3036--0.4008) & 0.2029 (0.1445--0.2530) & 0.2806 (0.2467--0.3025) \\
PR-AUC     & 0.3714 (0.2763--0.4145) & 0.2579 (0.0984--0.3852) & 0.2863 (0.2260--0.3689) \\
\bottomrule
\end{tabular}%
}
\end{table}

\subsection{Ensemble Performance}

\subsubsection{Balanced Testing Set}
On the balanced testing dataset ($n = 68$; 34 meningitis : 34 non-meningitis), the stacking ensemble showed an AUC of 0.9424 (95\% CI: 0.8840--0.9870), which is higher than the maximum cross-validated AUC of 0.8784 for the best performing base model. The sensitivity and specificity were 0.8514 (95\% CI: 0.7273--0.9655) and 0.9122 (95\% CI: 0.8064--1.0000), respectively. The PPV was 0.9066 (95\% CI: 0.7879--1.0000) and the NPV was 0.8601 (95\% CI: 0.7368--0.9688). The F1-score was 0.8764 (95\% CI: 0.7879--0.9552), and the MCC was 0.7651 (95\% CI: 0.6106--0.9120) while the PR-AUC was 0.9491 (95\% CI: 0.8906--0.9905). All results are provided in Table~\ref{tab:4}.

\begin{table}[htbp]
\caption{Stacking Ensemble Performance on the Balanced Testing Set. Metrics are reported as bootstrap mean (95\% CI, $B = 1{,}000$). Test set: $n = 68$ (34 meningitis : 34 non-meningitis). Decision threshold selected by Youden's J statistic.}
\label{tab:4}
\centering
\scriptsize
\setlength{\tabcolsep}{4pt}
\resizebox{0.85\columnwidth}{!}{%
\begin{tabular}{lc}
\toprule
\textbf{Metric} & \textbf{Ensemble (95\% CI)} \\
\midrule
AUC        & 0.9424 (0.8840--0.9870) \\
Sensitivity & 0.8514 (0.7273--0.9655) \\
Specificity & 0.9122 (0.8064--1.0000) \\
PPV        & 0.9066 (0.7879--1.0000) \\
NPV        & 0.8601 (0.7368--0.9688) \\
F1-score   & 0.8764 (0.7879--0.9552) \\
MCC        & 0.7651 (0.6106--0.9120) \\
PR-AUC     & 0.9491 (0.8906--0.9905) \\
\bottomrule
\end{tabular}%
}
\end{table}

\subsubsection{Imbalanced Testing Set}
\begin{figure}[htbp]
\centering
\includegraphics[width=\linewidth]{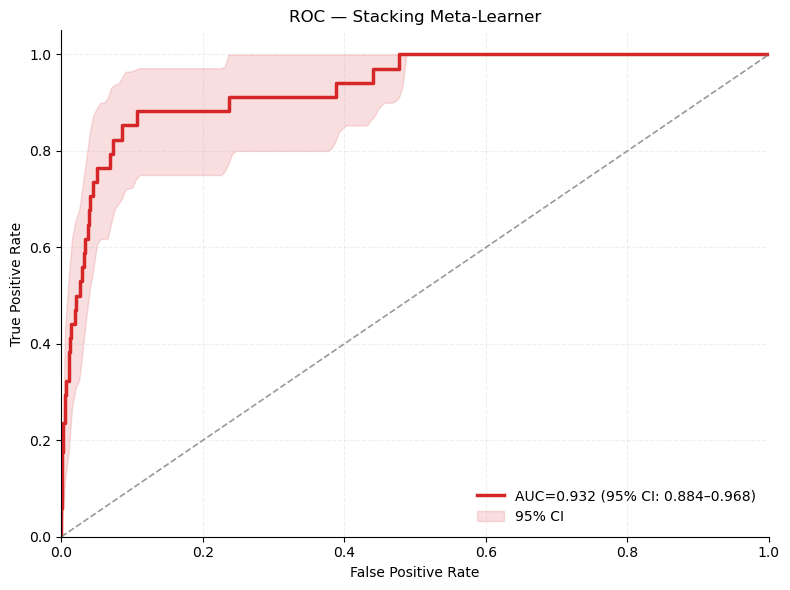}
\caption{Internal-test ROC curve for the ensemble model trained on 5,580 samples and evaluated on the imbalanced internal test set ($n = 7{,}382$). The curve summarizes ensemble discrimination under the clinically realistic class distribution used for the primary internal evaluation.}
\label{fig:3}
\end{figure}

On the imbalanced testing dataset ($n = 7{,}382$; 34 meningitis : 7,348 non-meningitis; 1:216 prevalence), the ensemble outputted an AUC of 0.9305 (95\% CI: 0.8820--0.9671) with the sensitivity and specificity of 0.8540 (95\% CI: 0.7222--0.9643) and 0.9122 (95\% CI: 0.9057--0.9187) respectively based on a Youden's J threshold of 0.5961. The NPV was 0.9993 (95\% CI: 0.9987--0.9999). The MCC was 0.1788 (95\% CI: 0.1365--0.2185) and the PR-AUC was 0.2053 (95\% CI: 0.0841--0.3398). Among the base models, the RF has the highest raw AUC of 0.9576 (95\% CI: 0.9327--0.9761) with a low Youden's J threshold of 0.0560 resulting in lower specificity (0.8671) and PPV (0.0315) compared to the ensemble. LightGBM and DNN achieved AUC values of 0.8249 and 0.8245 respectively. DeLong tests confirmed statistically  higher ensemble AUC over LightGBM ($z = 3.768$, $p = 0.0002$) and DNN ($z = 3.286$, $p = 0.0010$); however, no significant difference was found compared to RF ($z = -1.650$, $p = 0.0989$). The internal-test ROC curve is shown in Figure~\ref{fig:3}, with full results and DeLong statistics provided in Table~\ref{tab:5}.

\begin{table*}[t]
\caption{Model Performance on the Imbalanced Testing Set with DeLong Test Results. All four models evaluated on $n = 7{,}382$ (34 meningitis : 7,348 non-meningitis; 1:216 prevalence). Metrics reported as bootstrap mean (95\% CI, $B = 1{,}000$). Thresholds selected by Youden's J statistic per model. DeLong $p$-values compare each base model AUC against the ensemble AUC.}
\label{tab:5}
\centering
\scriptsize
\setlength{\tabcolsep}{4pt}
\resizebox{\textwidth}{!}{%
\begin{tabular}{@{}lcccc@{}}
\toprule
\textbf{Metric} & \textbf{RF (95\% CI)} & \textbf{LightGBM (95\% CI)} & \textbf{DNN (95\% CI)} & \textbf{Ensemble (95\% CI)} \\
\midrule
AUC        & 0.9576 (0.9327--0.9761) & 0.8249 (0.7424--0.9016) & 0.8245 (0.7145--0.9090) & 0.9305 (0.8820--0.9671) \\
Threshold  & 0.0560 & 0.4415 & 0.0118 & 0.5961 \\
Sensitivity & 0.9420 (0.8571--1.0000) & 0.6818 (0.5161--0.8387) & 0.6795 (0.5135--0.8235) & 0.8540 (0.7222--0.9643) \\
Specificity & 0.8671 (0.8596--0.8745) & 0.8598 (0.8521--0.8679) & 0.8832 (0.8763--0.8904) & 0.9122 (0.9057--0.9187) \\
PPV        & 0.0315 (0.0216--0.0433) & 0.0219 (0.0133--0.0310) & 0.0261 (0.0164--0.0384) & 0.0428 (0.0281--0.0594) \\
NPV        & 0.9997 (0.9992--1.0000) & 0.9983 (0.9972--0.9992) & 0.9983 (0.9972--0.9992) & 0.9993 (0.9987--0.9999) \\
F1-score   & 0.0610 (0.0421--0.0830) & 0.0423 (0.0261--0.0596) & 0.0502 (0.0318--0.0731) & 0.0814 (0.0543--0.1113) \\
MCC        & 0.1584 (0.1280--0.1900) & 0.1041 (0.0688--0.1370) & 0.1167 (0.0776--0.1568) & 0.1788 (0.1365--0.2185) \\
PR-AUC     & 0.1849 (0.0813--0.3171) & 0.0998 (0.0348--0.2116) & 0.2083 (0.0789--0.3521) & 0.2053 (0.0841--0.3398) \\
DeLong $p$ vs Ensemble & 0.0989 (ns) & 0.0002 (***) & 0.0010 (**) & --- \\
\bottomrule
\end{tabular}%
}
\end{table*}

\subsubsection{Meta-Learner Coefficients and Calibration}
Based on bootstrapping, the logistic regression meta-learner generated coefficients: $\beta_0 = -2.082$ (95\% CI: $-2.409$ to $-1.776$), $\beta_{\text{RF}} = +13.068$ (95\% CI: $+10.883$ to $+15.381$), $\beta_{\text{LGB}} = +3.173$ (95\% CI: $+2.471$ to $+3.769$), and $\beta_{\text{DNN}} = +1.149$ (95\% CI: $+0.506$ to $+1.959$). Meta-learner coefficients across all training configurations are provided in Supplementary Table~\ref{tab:s2}. For the imbalanced testing set, the Brier score of the ensemble was 0.1241, and the calibration curve and decision curve analysis are presented in Supplementary Figures~\ref{fig:s2} and~\ref{fig:s3}.

\subsection{External Validation}
\begin{figure}[htbp]
\centering
\includegraphics[width=\linewidth]{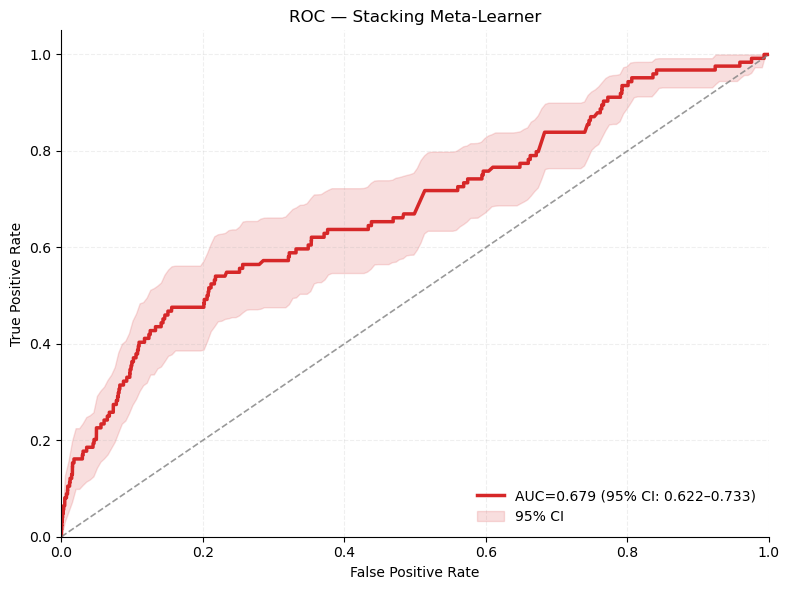}
\caption{The external-test ROC curve for the ensemble model trained on 5,580 samples and evaluated on the eICU validation cohort ($n = 27{,}156$). This curve summarizes generalization performance under the full external class distribution.}
\label{fig:external_roc}
\end{figure}

External validation of the model was performed on the eICU Collaborative Research Database \cite{ref44}, which consisted of 27,156 patients (124 meningitis cases and 27,032 non-meningitis; 1:218 prevalence) obtained from several different healthcare institutions. The approach of external validation followed previously established frameworks for external validation of clinical prediction models using large e-health record datasets \cite{ref38}. With the ICD-9 coding system (ICD-9 322.x), cases were classified in accordance with the MIMIC-III cohort definition. The ensemble performed an AUC of 0.6855 (95\% CI: 0.6290--0.7388), with a sensitivity of 0.8455 (95\% CI: 0.7786--0.9070) and a specificity of 0.2725 (95\% CI: 0.2671--0.2778). The NPV was 0.9974 (95\% CI: 0.9962--0.9985) and the PPV was low (0.0053; 95\% CI: 0.0043--0.0064) due to the 1:218 imbalance of the external cohort. The F1-score was 0.0105 (95\% CI: 0.0086--0.0126) and the MCC was 0.0178 (95\% CI: 0.0077--0.0272). The PR-AUC was 0.0367 (95\% CI: 0.0149--0.0689). The external validation ROC curve is presented in Figure~\ref{fig:external_roc}, with complete results shown in Table~\ref{tab:6}.

\begin{table}[htbp]
\caption{Stacking Ensemble Performance on the eICU External Validation Cohort.}
\label{tab:6}
\centering
\scriptsize
\setlength{\tabcolsep}{4pt}
\resizebox{0.85\columnwidth}{!}{%
\begin{tabular}{lc}
\toprule
\textbf{Metric} & \textbf{Ensemble (95\% CI)} \\
\midrule
AUC        & 0.6855 (0.6290--0.7388) \\
Sensitivity & 0.8455 (0.7786--0.9070) \\
Specificity & 0.2725 (0.2671--0.2778) \\
PPV        & 0.0053 (0.0043--0.0064) \\
NPV        & 0.9974 (0.9962--0.9985) \\
F1-score   & 0.0105 (0.0086--0.0126) \\
MCC        & 0.0178 (0.0077--0.0272) \\
PR-AUC     & 0.0367 (0.0149--0.0689) \\
\bottomrule
\end{tabular}%
}
\end{table}

\subsection{SHAP-Based Interpretability}
\begin{figure*}[t]
\centering
\includegraphics[width=0.98\textwidth]{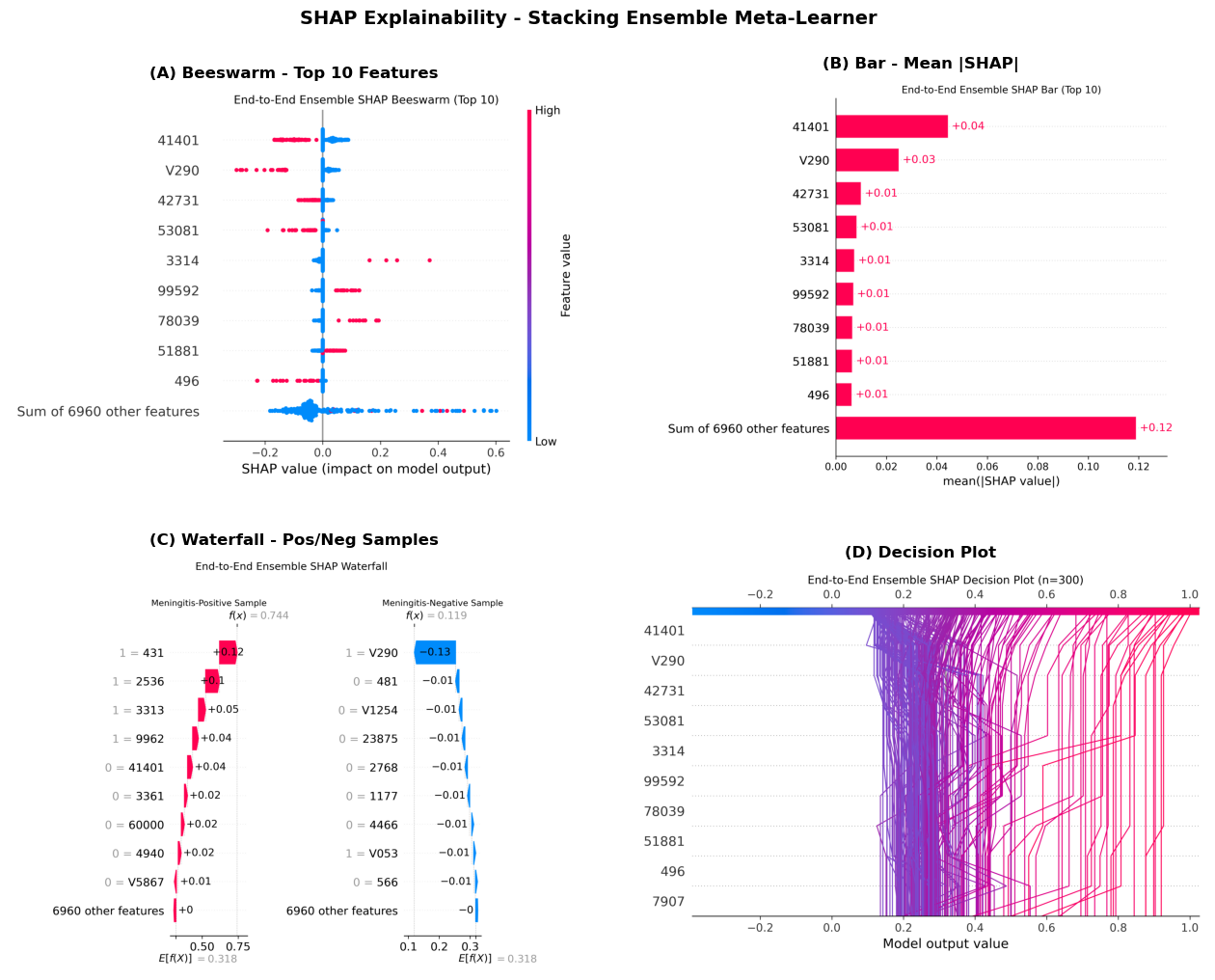}
\caption{SHAP Explainability Panel for the Stacking Ensemble Meta-Learner. Four-panel figure: (A) Beeswarm plot showing feature-level SHAP value distributions across all explained samples; (B) Bar plot of mean $|\phi|$ for top 10 features; (C) Waterfall plots for one meningitis-positive and one meningitis-negative representative sample; (D) Decision plot showing cumulative SHAP contributions across all 300 explained samples. Top 10 features shown in all panels.}
\label{fig:4}
\end{figure*}

SHAP's KernelExplainer was applied to the meta-learner to quantify the contributions of individual features across all parts of the ensemble; the reference set used for this analysis consisted of 50 samples, and the explanation set consisted of 300 samples from the imbalanced testing set. The ten features that had the highest mean SHAP absolute values (mean $|\phi|$) are shown in Table~\ref{tab:7}. The highest ranked feature was coronary atherosclerosis (41401; mean $|\phi| = 0.0459$), followed by observation for a suspected infectious condition (V290; mean $|\phi| = 0.0277$) and atrial fibrillation (42731; mean $|\phi| = 0.0115$). The SHAP ranking of the features differs from the RF ranking in Table~\ref{tab:2}. The beeswarm, bar, waterfall, and decision plots are shown in Figure~\ref{fig:4} \cite{ref34,ref36}.

\begin{table}[htbp]
\caption{
Top 10 features with the highest mean absolute values; Positive $\phi$ values increase the predicted meningitis probability while negative $\phi$ values decrease the predicted meningitis probabilities.
}
\label{tab:7}
\centering
\scriptsize
\setlength{\tabcolsep}{3pt}
\resizebox{\columnwidth}{!}{%
\begin{tabular}{@{}clp{0.48\columnwidth}c@{}}
\toprule
\textbf{Rank} & \textbf{ICD-9 Code} & \textbf{Clinical Description} & \textbf{Mean $|\phi|$} \\
\midrule
1  & 41401 & Coronary atherosclerosis of native coronary artery     & 0.0459 \\
2  & V290  & Observation for suspected infectious condition         & 0.0277 \\
3  & 42731 & Atrial fibrillation                                    & 0.0115 \\
4  & 53081 & Esophageal reflux                                      & 0.0084 \\
5  & 99592 & Severe sepsis                                          & 0.0078 \\
6  & 3314  & Obstructive hydrocephalus                              & 0.0073 \\
7  & 496   & Chronic airway obstruction, NEC                        & 0.0064 \\
8  & 78039 & Other alteration of consciousness                      & 0.0061 \\
9  & 51881 & Acute respiratory failure                              & 0.0059 \\
10 & 7907  & Bacteremia                                             & 0.0058 \\
\bottomrule
\end{tabular}%
}
\end{table}

\section{Discussion}
The research in this paper demonstrated that a stacking ensemble learning technique including Random Forest, LightGBM, and DNN produced robust results for identifying meningitis using EHR datasets, as evaluated by the DeLong method on both the imbalanced internal test set and an external cohort. The model takes advantage of each base model's unique interpretation of the data and learns the patterns generated by the base models. While this research mainly leveraged ICD-9 codes and gender, it indicated that a substantial proportion of clinically relevant variability can be captured by these features. Limitations include the relatively small number of meningitis samples and the absence of laboratory or vital sign data. Future research should incorporate more relevant medical features and include additional sites for validation.
\section{Conclusions}

By leveraging coded administrative data from EHRs, the stacking ensemble of RF, LightGBM, and a DNN is potentially capable of detecting EHR-based meningitis. The performance of the model on the internal test data was impressive and demonstrated at least equivalent sensitivity in an independent external cohort, supporting it as a potential rule-out tool for meningitis detection in the ER/ICU settings.

\section*{Declaration of Interests}
The authors declare that they have no known competing financial interests or personal relationships that could have appeared to influence the work reported in this paper.

\clearpage
\bibliographystyle{elsarticle-num}
\bibliography{references}

\clearpage
\newgeometry{top=30mm,bottom=30mm}
\setcounter{table}{0}
\renewcommand{\thetable}{S\arabic{table}}
\setcounter{figure}{0}
\renewcommand{\thefigure}{S\arabic{figure}}
\makeatletter
\renewcommand{\theHtable}{stable.\arabic{table}}
\renewcommand{\theHfigure}{sfigure.\arabic{figure}}
\makeatother
\onecolumn
\section*{Supplementary Material}

\subsection*{1. Supplementary Figure S1. Base-Model Cross-Validation ROC Curves}

\begin{figure}[htbp]
\centering
\begin{minipage}[t]{0.31\linewidth}
\centering
\includegraphics[width=\linewidth]{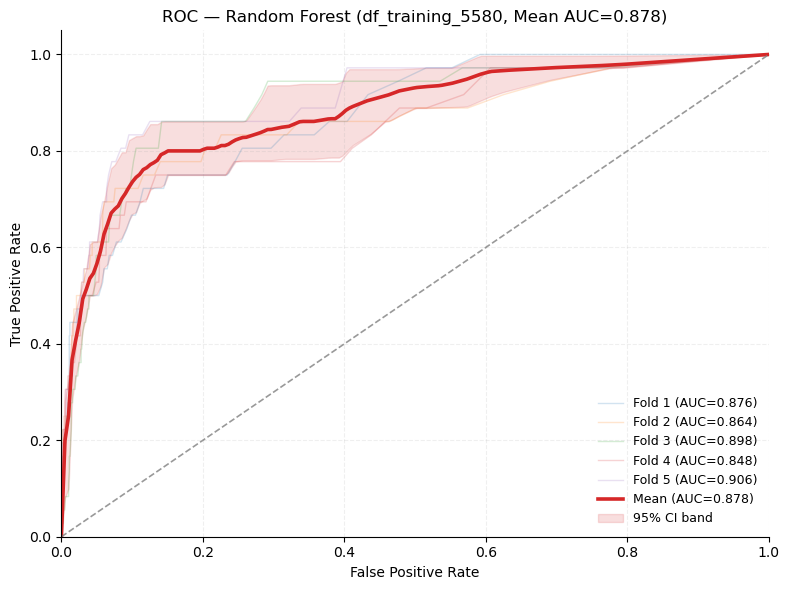}
\end{minipage}\hfill
\begin{minipage}[t]{0.31\linewidth}
\centering
\includegraphics[width=\linewidth]{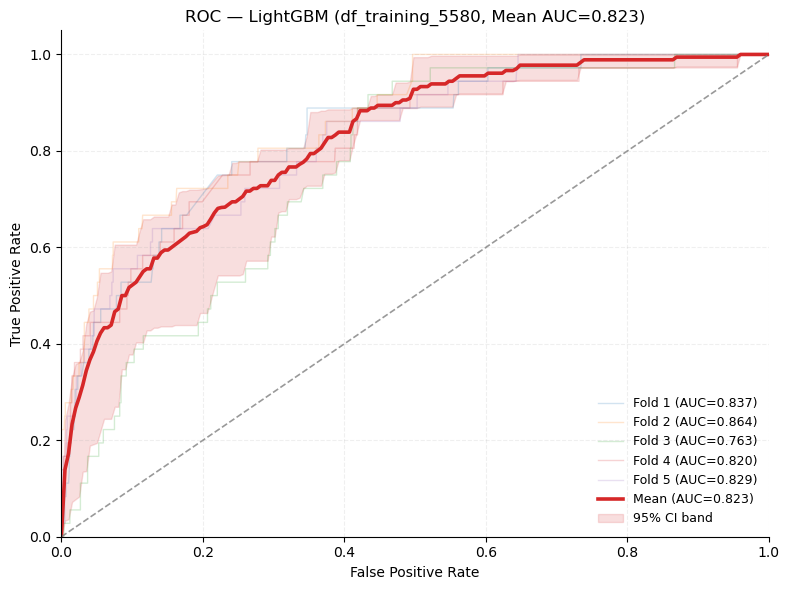}
\end{minipage}\hfill
\begin{minipage}[t]{0.31\linewidth}
\centering
\includegraphics[width=\linewidth]{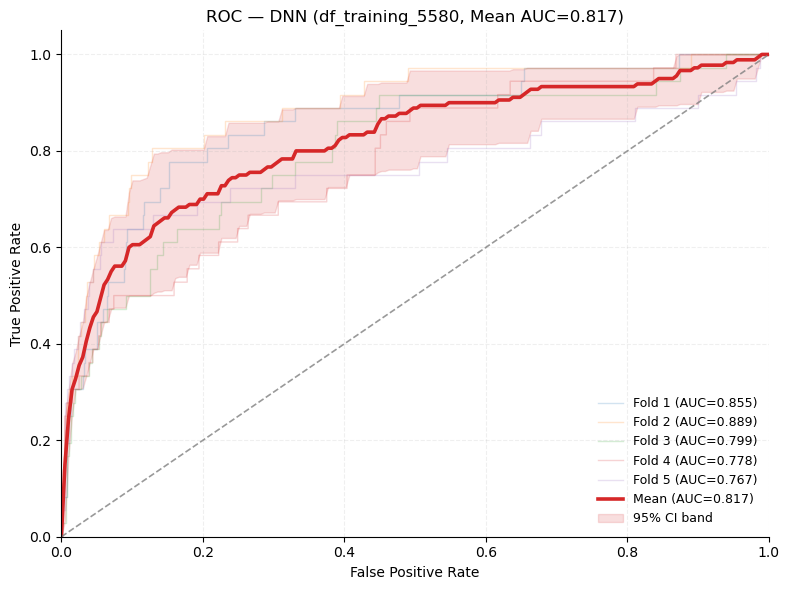}
\end{minipage}
\caption{Base-model cross-validation ROC curves for the primary training configuration (Train = 5,580). Left: Random Forest. Middle: LightGBM. Right: Deep Neural Network.}
\label{fig:s1}
\end{figure}

\subsection*{2. Supplementary Figures S2 and S3. Calibration Curve and Decision Curve Analysis}

\begin{figure}[htbp]
\centering
\begin{minipage}[t]{0.48\linewidth}
\centering
\includegraphics[width=\linewidth]{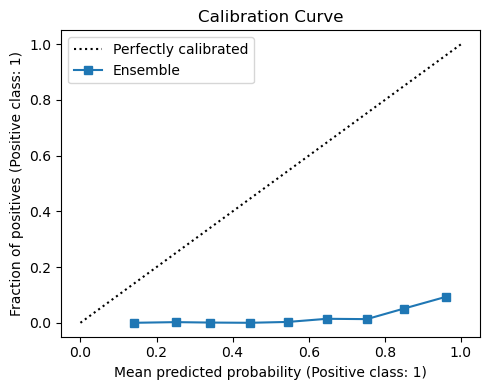}
\par\vspace{0.5em}
\refstepcounter{figure}\label{fig:s2}
\textbf{Figure~\thefigure.} Calibration Curve for the stacking ensemble on the imbalanced internal test set ($n = 7{,}382$), computed using 10 equal-width probability bins. The Brier score is 0.1241. 
\end{minipage}\hfill
\begin{minipage}[t]{0.48\linewidth}
\centering
\includegraphics[width=\linewidth,height=0.365\textheight,keepaspectratio]{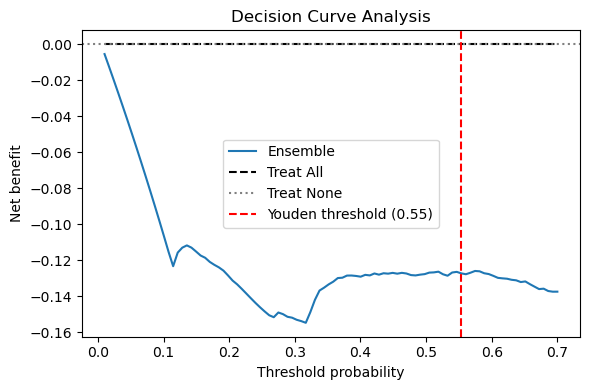}
\par\vspace{0.5em}
\refstepcounter{figure}\label{fig:s3}
\textbf{Figure~\thefigure.} Decision Curve Analysis. Net clinical benefit vs.\ decision threshold for the stacking ensemble, RF-only, LightGBM-only, and DNN-only on the imbalanced internal test set.
\end{minipage}
\end{figure}

\subsection*{3. Supplementary Table S1. Ensemble Performance Across All eICU Control Subset Sizes}

Ensemble performance on four eICU validation cohorts varying the number of randomly sampled non-meningitis controls (threshold = 0.3; 124 meningitis cases fixed). All 95\% CIs: 1,000 bootstrap resamples.

\begin{table}[htbp]
\caption{Ensemble Performance Across All eICU Control Subset Sizes.}
\label{tab:s1}
\centering
\footnotesize
\setlength{\tabcolsep}{4pt}
\resizebox{0.96\textwidth}{!}{%
\begin{tabular}{@{}lclcccccccc@{}}
\toprule
\textbf{Cohort} & \textbf{Thresh.} & \textbf{$n$ (Men./Control)} & \textbf{AUC (95\% CI)} & \textbf{Sensitivity (95\% CI)} & \textbf{Spec.} & \textbf{PPV} & \textbf{NPV} & \textbf{F1} & \textbf{MCC} & \textbf{PR-AUC} \\
\midrule
val\_1364  & 0.2 & 124 / 1,240  & 0.6777 (0.6235--0.7310) & 0.9678 (0.9338--0.9925) & 0.1013 & 0.0973 & 0.9693 & 0.1767 & 0.0677 & 0.2832 \\
val\_2604  & 0.2 & 124 / 2,480  & 0.6854 (0.6356--0.7369) & 0.9688 (0.9348--1.0000) & 0.1056 & 0.0516 & 0.9854 & 0.0979 & 0.0524 & 0.2082 \\
val\_3844  & 0.3 & 124 / 3,720  & 0.6835 (0.6309--0.7319) & 0.9277 (0.8780--0.9685) & 0.1998 & 0.0373 & 0.9881 & 0.0716 & 0.0568 & 0.1555 \\
\textbf{val\_27156} & \textbf{0.3} & \textbf{124 / 27,032} & \textbf{0.6855 (0.6290--0.7388)} & \textbf{0.8455 (0.7786--0.9070)} & \textbf{0.2725} & \textbf{0.0053} & \textbf{0.9974} & \textbf{0.0105} & \textbf{0.0178} & \textbf{0.0367} \\
\bottomrule
\end{tabular}%
}
\end{table}

\noindent\textit{val\_27156 is the primary external validation cohort reported in Table~\ref{tab:5} of the main text.}

\subsection*{4. Supplementary Table S2. Meta-Learner Coefficients Across Training Sizes}

Coefficients of the stacking meta-learner (logistic regression) trained on out-of-fold predictions from 5-fold cross-validation, for all four training configurations. Primary configuration (Train = 5,580) used for all main-text results.

\begin{table}[htbp]
\caption{Meta-Learner Coefficients Across Training Sizes. Bootstrap 95\% confidence intervals were computed with $B = 1{,}000$ resamples of the meta-training feature matrix.}
\label{tab:s2}
\centering
\footnotesize
\setlength{\tabcolsep}{4pt}
\resizebox{0.96\textwidth}{!}{%
\begin{tabular}{@{}lcccc@{}}
\toprule
\textbf{Train Size} & \textbf{$\beta_0$ (95\% CI)} & \textbf{$\beta_{\text{RF}}$ (95\% CI)} & \textbf{$\beta_{\text{LGB}}$ (95\% CI)} & \textbf{$\beta_{\text{DNN}}$ (95\% CI)} \\
\midrule
360   & $-2.330$ ($-2.744$ to $-1.969$) & $+3.771$ ($+3.055$ to $+4.611$)    & $+0.038$ ($-0.495$ to $+0.672$)  & $+1.380$ ($+0.846$ to $+2.043$) \\
1,980 & $-2.387$ ($-2.735$ to $-1.993$) & $+8.863$ ($+7.483$ to $+10.286$)   & $+2.351$ ($+1.577$ to $+3.073$)  & $+0.952$ ($+0.450$ to $+1.450$) \\
3,780 & $-2.270$ ($-2.666$ to $-1.971$) & $+10.433$ ($+8.512$ to $+12.180$)  & $+3.167$ ($+2.500$ to $+3.968$)  & $+1.266$ ($+0.579$ to $+1.912$) \\
\textbf{5,580} & $\mathbf{-2.082}$ ($\mathbf{-2.409}$ to $\mathbf{-1.776}$) & $\mathbf{+13.068}$ ($\mathbf{+10.883}$ to $\mathbf{+15.381}$) & $\mathbf{+3.173}$ ($\mathbf{+2.471}$ to $\mathbf{+3.769}$) & $\mathbf{+1.149}$ ($\mathbf{+0.506}$ to $\mathbf{+1.959}$) \\
\bottomrule
\end{tabular}%
}
\end{table}

\subsection*{5. Supplementary Note 1. Temporal Filtering Procedure}

For each meningitis patient, ICD-9 diagnostic codes were filtered to retain only those with chart timestamps strictly prior to the first occurrence of any ICD-9 322.x code. Codes assigned on or after the meningitis diagnosis date were excluded. This procedure was applied before feature matrix construction and before any train/test split.

\restoregeometry

\end{document}